\documentclass[table]{ai2style/ai2}

\usepackage{microtype}
\usepackage{hyperref}
\usepackage{url}
\usepackage{booktabs, tabularx} %
\usepackage{graphicx}
\usepackage{lineno}
\usepackage{enumitem}
\usepackage{listings} %
\usepackage{svg}

\definecolor{ darkblue}{rgb}{0, 0, 0.5}
\hypersetup{colorlinks=true, citecolor=darkblue, linkcolor=darkblue, urlcolor=darkblue}

\usepackage{amssymb}
\usepackage{multirow}
\usepackage{bigdelim}
\usepackage{todonotes}
\usepackage{longtable}
\usepackage{tabularray}
\usepackage{wrapfig}
\usepackage[most]{tcolorbox}
\usepackage{url}
\usepackage{xspace}
\usepackage{svg}
\usepackage[absolute]{textpos} %

\usepackage{fdsymbol}   %

\usepackage[utf8]{inputenc} %
\usepackage[T1]{fontenc}    %
\usepackage{url}            %
\usepackage{booktabs}       %
\usepackage{amsfonts}       %
\usepackage{nicefrac}       %
\usepackage{microtype}      %
\usepackage[table,xcdraw]{xcolor}         %
\usepackage{amsmath}
\usepackage[most]{tcolorbox}
\usepackage{csquotes}
\usepackage{tablefootnote}

\usepackage{siunitx}
\usepackage{graphicx}
\usepackage{arydshln}
\usepackage{wrapfig}
\usepackage{enumitem}
\usepackage{soul} %

\definecolor{baselinecolor}{gray}{.9}

\usepackage{multirow}
\usepackage{graphicx}
\usepackage{listings}
\lstdefinelanguage{Prompt}{
  morestring=[b]",
}
\definecolor{codebg}{RGB}{245,245,245}
\usepackage{multirow}
\usepackage{xspace}
\usepackage{adjustbox}
\usepackage{pifont}
\usepackage{caption}
\usepackage{makecell}
\usepackage{subcaption}
\usepackage{bold-extra}
\usepackage{url}

\usepackage{pgf-pie}

\usepackage{hyperref}
\definecolor{linkcolor}{RGB}{0, 0, 128}
\hypersetup{
     colorlinks   = true,
     citecolor    = linkcolor,
     linkcolor    = linkcolor,
     urlcolor     = linkcolor,
}
\usepackage{pifont}%
\newcommand{\cmark}{\ding{51}}%
\newcommand{\xmark}{\ding{55}}%
\usepackage{listings}

\setlist[itemize]{leftmargin=*,itemsep=0em,parsep=0.3em,topsep=0.3em}

\DeclareUnicodeCharacter{2212}{\ensuremath{-}}

\definecolor{maroon}{HTML}{F26035}
\definecolor{yellow}{HTML}{FDBC42}
\definecolor{lavender}{HTML}{734f96}
\definecolor{darkergrey}{HTML}{444444}
\definecolor{midgrey}{HTML}{e6eded}
\definecolor{ai2pink}{HTML}{f0529c}%
\definecolor{ai2midpink}{HTML}{fad3e5}
\definecolor{ai2lightpink}{HTML}{fbecf3}
\definecolor{ai2midwhite}{HTML}{f2e5d9}
\definecolor{ai2offwhite}{HTML}{fbf4ee}
\definecolor{ai2green}{HTML}{0fcb8c}
\definecolor{ai2lightgreen}{HTML}{e7f9f3}
\definecolor{ai2darkgreen}{HTML}{105257}
\definecolor{ai2purple}{HTML}{B932EB}
\definecolor{ai2lightpurple}{HTML}{f7e8fc}
\definecolor{neutralEight}{HTML}{343434}
\definecolor{neutralFive}{HTML}{838383}
\definecolor{neutralThree}{HTML}{bebebe}
\definecolor{neutralOne}{HTML}{dedede}
\definecolor{lightgrey}{HTML}{fafcfc}
\definecolor{plum}{rgb}{0.56,0.27,0.52}

\usepackage{tikz}

\definecolor{maroon}{HTML}{F26035}
\definecolor{yellow}{HTML}{FDBC42}
\definecolor{darkred}{RGB}{156, 39, 33}
\definecolor{darkblue}{RGB}{31, 90, 153}
\definecolor{forestgreen}{rgb}{0.13, 0.55, 0.13}
\definecolor{brickred}{rgb}{0.8, 0.25, 0.33}
\definecolor{olmoDarkBlue}{HTML}{012e59}
\definecolor{olmoBlue}{HTML}{265ed4}
\definecolor{olmoLightBlue}{HTML}{012e59}
\definecolor{olmoTeal}{HTML}{00d5ff}
\definecolor{olmoYellow}{HTML}{ffbb00}
\definecolor{olmoOrange}{HTML}{ff9100}

\newcommand{\app}{\raise.17ex\hbox{$\scriptstyle\sim$}}
\renewcommand{\paragraph}[1]{\vspace{0.5mm}\noindent\textbf{#1}}

\usepackage{algorithm}
\usepackage{algpseudocode}
\usepackage[table]{xcolor}

\usepackage{pifont}

\definecolor{molmocolor}{RGB}{240, 82, 156}
\definecolor{tablegray}{RGB}{223, 242, 252}
\definecolor{tablegreen}{RGB}{15, 203, 150}
\definecolor{tableyellow}{RGB}{250, 242, 233}
\definecolor{tableblue}{RGB}{240, 82, 156}
\definecolor{darkpink}{RGB}{139, 14, 98}

\usepackage{setspace}

\usepackage{nicematrix}
\newcolumntype{L}[1]{>{\raggedright\let\newline\\\arraybackslash\hspace{0pt}}m{#1}}
\newcolumntype{C}[1]{>{\centering\let\newline\\\arraybackslash\hspace{0pt}}m{#1}}
\newcolumntype{R}[1]{>{\raggedleft\let\newline\\\arraybackslash\hspace{0pt}}m{#1}}
\newcolumntype{P}[1]{>{\centering\let\newline\\\arraybackslash\columncolor{ai2lightpink}}m{#1}}
\newcommand{\github}{\raisebox{-1.5pt}{\includegraphics[height=1.05em]{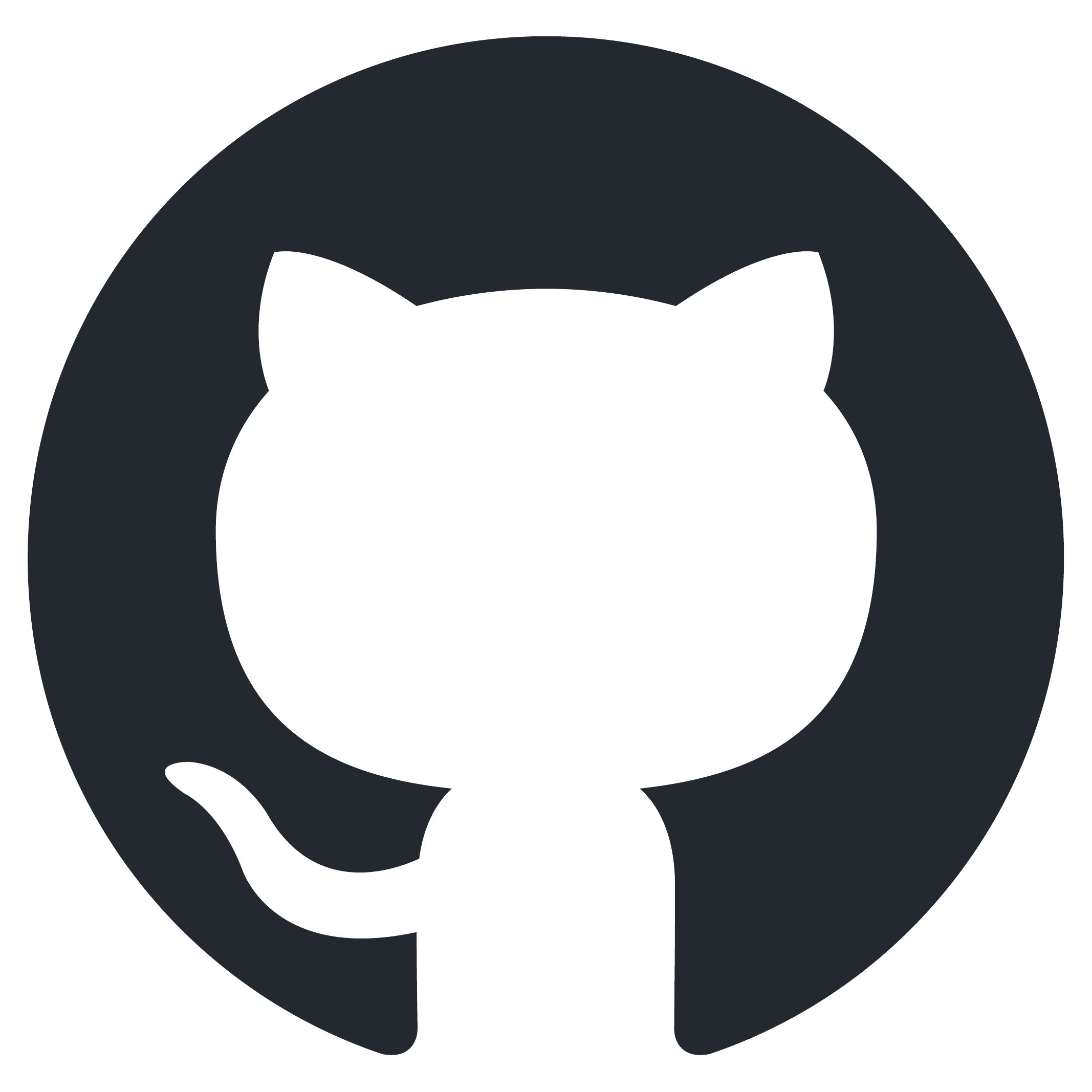}}\xspace}

\makeatletter
\DeclareRobustCommand\onedot{\futurelet\@let@token\@onedot}
\def\@onedot{\ifx\@let@token.\else.\null\fi\xspace}

\makeatother

\title{Unified Spatio-Temporal Token Scoring for Efficient Video VLMs}

\authorOne[1]{Jianrui Zhang*}
\authorOne[2]{Yue Yang}
\authorOne[2]{Rohun Tripathi}
\authorOne[2]{Winson Han}
\authorTwo[2]{Ranjay Krishna}
\authorTwo[2]{Christopher Clark$^\dagger$}
\authorTwo[1]{Yong Jae Lee$^\dagger$}
\authorTwo[2]{Sangho Lee$^\dagger$}

\affiliation[1]{University of Wisconsin-Madison}
\affiliation[2]{Allen Institute for AI}

\contribution[*]{ Work done during Jianrui (Harris)'s internship at Ai2, $^\dagger$ denotes equal advising.}

\definecolor{molmocolor}{RGB}{240, 82, 156}
\definecolor{tablegray}{RGB}{223, 242, 252}
\definecolor{tablegreen}{RGB}{15, 203, 150}
\definecolor{tableyellow}{RGB}{250, 242, 233}
\definecolor{tableblue}{RGB}{240, 82, 156}
\definecolor{darkpink}{RGB}{139, 14, 98}
\definecolor{baselinecolor}{gray}{.9}

\usepackage{graphicx}
\usepackage{booktabs}
\usepackage{multirow}
\usepackage{caption}
\usepackage{wrapfig}

\newcommand{\mycell}[2]{%

  \rotatebox{90}{%
    \parbox{2.0cm}{%
      \setlength{\baselineskip}{0.5em}%
      \tiny\textbf{#1}\\
      \footnotesize{\tiny\textcolor{gray}{#2}}%
    }%
  }%
}

\newcommand{\newcell}[1]{%
  \rotatebox{90}{%
    \parbox{2.0cm}{%
      \setlength{\baselineskip}{0.5em}%
      \tiny\textbf{#1}
    }%
  }%
}

\abstract{
Token pruning is essential for enhancing the computational efficiency of vision-language models (VLMs), particularly for video-based tasks where temporal redundancy is prevalent.
Prior approaches typically prune tokens either (1) within the vision transformer (ViT) exclusively for unimodal perception tasks such as action recognition and object segmentation, without adapting to downstream vision-language tasks; or (2) only within the LLM while leaving the ViT output intact, often requiring complex text-conditioned token selection mechanisms.
In this paper, we introduce \textbf{S}patio-\textbf{T}emporal \textbf{T}oken \textbf{S}coring (\textbf{STTS}), a simple and lightweight module that prunes vision tokens across both the ViT and the LLM without text conditioning or token merging, and is fully compatible with end-to-end training. By learning how to score temporally via an auxiliary loss and spatially via LLM downstream gradients, aided by our efficient packing algorithm, STTS prunes 50\% of vision tokens throughout the entire architecture, resulting in a 62\% improvement in efficiency during both training and inference with only a 0.7\% drop in average performance across 13 short and long video QA tasks. Efficiency gains increase with more sampled frames per video. Applying test-time scaling for long-video QA further yields performance gains of 0.5-1\% compared to the baseline.
Overall, STTS represents a novel, simple yet effective technique for unified, architecture-wide vision token pruning.

}

\metadata[\vspace{.5em}\quad\github Code:]{
    \href{https://github.com/allenai/STTS}{\texttt{https://github.com/allenai/STTS}} \quad
}

\begin{document}

\maketitle

\section{Introduction}
\label{sec:intro}
The rapid progress of vision-language models (VLMs) in video understanding has come at a substantial computational cost. Processing video requires encoding a large number of frames, each decomposed into hundreds of patch tokens by a vision transformer (ViT)~\cite{dosovitskiy2021vit}. As the number of frames increases, the resulting token sequences become quadratically expensive under attention, leading to significant memory usage, reduced training throughput, and increased inference latency. This long visual token sequence not only burdens the ViT encoder but also amplifies the computational load of the large language model (LLM) that consumes its output. Token pruning, selectively discarding uninformative visual tokens, offers a natural solution and has attracted considerable research attention in recent years.

Existing pruning methods, however, address only part of the problem. Pre-ViT and in-ViT approaches reduce token redundancy before or during ViT encoding, employing strategies such as early exiting~\cite{tang2023dynamictokenpruningplain}, token matching and mixing~\cite{vasu2023fastvitfasthybridvision,bolya2023tokenmergingvitfaster}, and attention-based scoring~\cite{kong2022spvitenablingfastervision,chen2025vltp}. While effective for spatial redundancy in unimodal perception tasks, these methods are not explicitly designed for multimodal VLM objectives and do not account for cross-frame temporal redundancy in video inputs. Post-ViT approaches, on the other hand, prune the tokens passed from the ViT to the LLM via spatial pooling~\cite{shang2025prumerge, wu2024freevaofflinemllmtrainingfree, cai2024matryoshkamultimodalmodels, hu2024matryoshkaquerytransformerlarge}, text-conditioned selection~\cite{huang2024prunevidvisualtokenpruning, liu2025videoxlproreconstructivetokencompression, luo2025vcmvisionconceptmodeling}, or cross-frame merging~\cite{choudhury2024rlt, hyun2025multigranularspatiotemporaltokenmerging, shao2025holitomholistictokenmerging, shen2025fastviddynamicdensitypruning}. These methods, however, leave the ViT encoder untouched, even though the ViT constitutes a major computational bottleneck for video inputs, as its cost grows linearly with the number of frames. Neither paradigm, therefore, provides a holistic solution for scalable video VLMs.

\begin{figure*}[t]
	\begin{center}
	\centering
        \includegraphics[width=\linewidth]{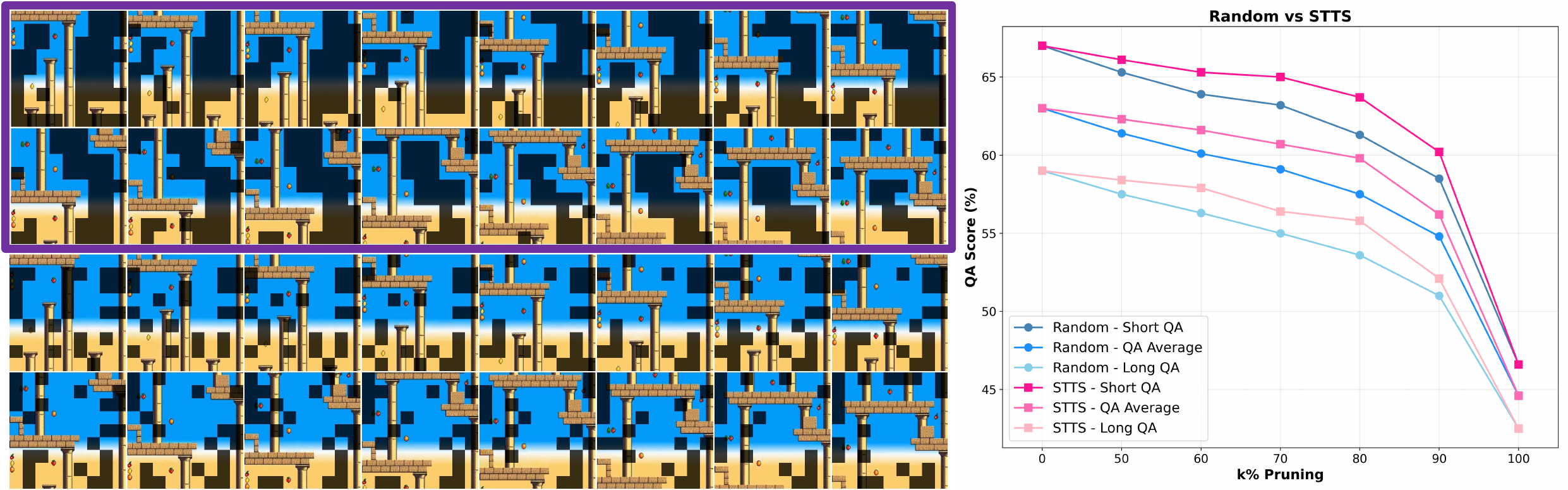}
        \caption{(Left) Token pruning with our STTS (purple box) vs. a cosine-similarity-based heuristic. STTS learns that background patches are less important, while the heuristic prunes all tokens equally. (Right) QA performance under increasing vision token pruning ratios ($k\%$). STTS (pink squares) consistently demonstrates a flatter, more robust degradation curve compared to the Random baseline (blue circles) across all metrics.}
		\label{fig:teaser}
	\end{center}
\end{figure*}

We introduce \emph{Spatio-Temporal Token Scoring} (\textbf{STTS}), a light-weight module designed to seamlessly bridge the gap between performance and efficiency in video understanding. Rather than relying on cumbersome architectural changes or complex and expensive token selection algorithms, STTS provides a streamlined, end-to-end trainable solution that directly reduces the visual token burden across the entire VLM pipeline. 
This accelerates both training and inference phases without sacrificing the model's fundamental reasoning capabilities.

In this paper, we will demonstrate the efficiency and effectiveness of STTS, and our main contributions are fourfold:
\begin{enumerate}
    \item \textbf{Unified Token Pruning Module:} STTS is a lightweight, end-to-end trainable module that seamlessly prunes visual tokens across both the ViT and LLM without requiring significant architectural modifications, text-conditioned selection, or complex merging algorithms.
    \item \textbf{Dual-Axis Scoring Mechanism:} STTS scores tokens by simultaneously targeting intra-frame spatial saliency learned implicitly via downstream multimodal objectives (Figure~\ref{fig:teaser}) and inter-frame temporal redundancy regularized through an auxiliary loss.
    \item \textbf{Significant Efficiency Gains:} STTS can safely drop 50\% of visual tokens, improving both training throughput and inference efficiency by up to 62\% with negligible performance loss.
    \item \textbf{Scalibility:} Sampling more video frames further increases efficiency gains. Additionally, we show its generalizability by applying test-time scaling to achieve consistent 0.5--1\% improvements on long-video QA benchmarks.
\end{enumerate}

\section{Related Works}
\subsection{Pre-/In-ViT Token Pruning}

A significant body of work has focused on token pruning and merging within Image-ViTs. For example, SPViT~\cite{kong2022spvitenablingfastervision} aggregates redundant tokens into a single 'package token', while FastViT~\cite{vasu2023fastvitfasthybridvision} and ToMe~\cite{bolya2023tokenmergingvitfaster} employ token mixing and matching, respectively, to efficiently merge tokens. These methods, however, primarily focus on spatial pruning within static images and do not address the temporal redundancies inherent in video.

Other approaches focus on different pruning criteria. DToP~\cite{tang2023dynamictokenpruningplain} uses early-exiting to stop processing ``easy'' tokens for instance segmentation. VLTP~\cite{chen2025vltp} employs a pruning decoder to select important tokens at specific ViT layers. Run-Length Tokenization~\cite{choudhury2024rlt} identifies temporally redundant patches even before they enter the ViT. However, these techniques are typically demonstrated on vision-only tasks like segmentation or action classification and have not been extended to downstream VLM, and specifically video-LLM, applications.

In contrast to these works, STTS is designed as a simple, merge-free module that prunes both spatially and temporally within the ViT and is explicitly evaluated on downstream video-LLM tasks.

\subsection{Post-ViT Vision Token Pruning}

Another line of research focuses on pruning vision tokens exclusively post-ViT—that is, between the vision encoder and the LLM. For instance, FreeVA~\cite{wu2024freevaofflinemllmtrainingfree} provides a training-free method for temporal token aggregation. PruneVid~\cite{huang2024prunevidvisualtokenpruning}, STTM~\cite{hyun2025multigranularspatiotemporaltokenmerging}, and HoliTom~\cite{shao2025holitomholistictokenmerging} merge tokens both spatially and temporally before they are fed to the LLM. FastVid~\cite{shen2025fastviddynamicdensitypruning} incorporates temporal segmentation to guide its merging process. Similarly, LLaVA-PruMerge~\cite{shang2025prumerge} leverages CLIP-ViT attention scores for merging. More complex methods like VCM~\cite{luo2025vcmvisionconceptmodeling} and Video-XL-Pro~\cite{liu2025videoxlproreconstructivetokencompression} employ query-based selector modules that require cross-attention with text tokens. \cite{cai2024matryoshkamultimodalmodels, hu2024matryoshkaquerytransformerlarge} utilizes Matryoshka representations to compress vision tokens into different levels of granularity.

A critical limitation of all these methods is that they prune \textbf{after} the ViT. Consequently, the ViT must still process every frame from the input video, creating a significant computational bottleneck, especially for long inputs. Furthermore, many of these approaches rely on complex merging algorithms or text-conditioned modules. STTS addresses both limitations by applying a simple, merge-free scoring mechanism that prunes starting in the ViT and thus naturally reduces the compute needed in the LLM.

\section{Spatio-Temporal Token Scoring (STTS)}

\begin{wrapfigure}{l}{0.5\textwidth}
\vspace{-2em}
    \centering
    \includegraphics[width=\linewidth]{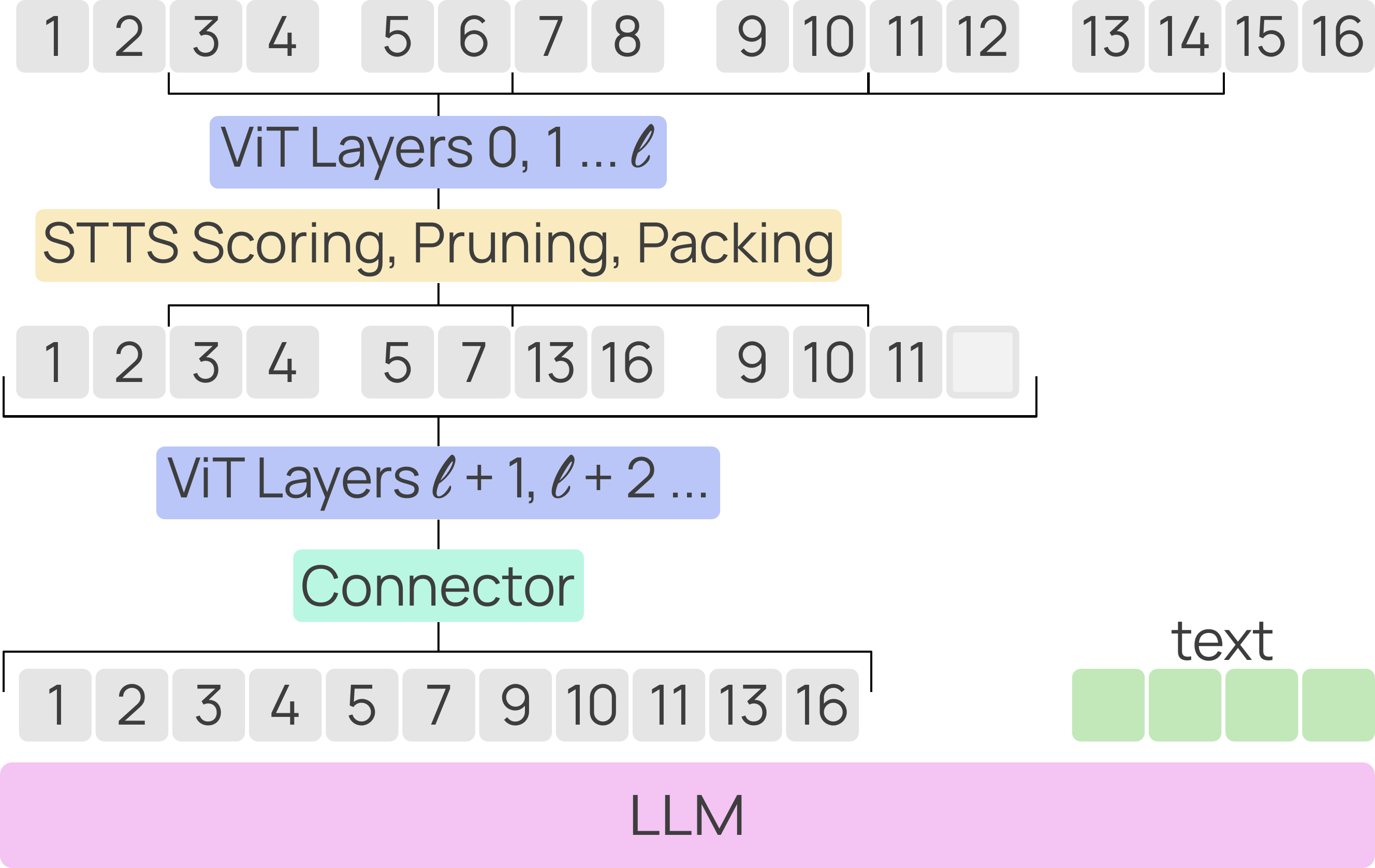}
    \caption{Overall workflow of using STTS within the VLM. Numbered vision tokens here are 3x3 grids. After ViT layer $l$, STTS prunes vision tokens permanently from the entire architecture. We pad tokens during packing for ViT batch computation.}
    \label{fig:overview}
    \vspace{-1em}
\end{wrapfigure}

The goal of our paper is to minimize compute spent on vision tokens as much as possible without significantly damaging the model's video reasoning capabilities. Formally, we frame this as a constrained optimization objective. Let $N_{\text{total}} = T \times N$ be the total number of initial patch tokens across all frames. We seek to find the optimal model parameters $\theta$ that minimize the overall loss $\mathcal{L}$, subject to a strict computational budget defined by our pruning ratio $k$:

\begin{equation*}
\min_{\theta} \mathcal{L}(\theta): \|\mathcal{M}\|_0 \leq \left(1 - k\%\right) N_{\text{total}}
\end{equation*}

where $\mathcal{M} \in \{0, 1\}^{T \times N}$ is a binary mask representing the retained tokens after scoring and $\mathcal{L}$ encompasses both the primary VLM reasoning task and our temporal auxiliary loss (detailed in Section~\ref{sec: stts aux loss}).

Figure~\ref{fig:overview} illustrates the overall architecture of our framework. Our model follows the common design of modern VLMs, combining a pre-trained LLM with a ViT~\cite{dosovitskiy2021vit} via a connector module~\cite{molmov1,liu2023llava}. Concretely, we build upon Molmo2~\cite{clark2026molmo2openweightsdata} as our backbone, which applies $w \times w$ spatial pooling ($w=3$ by default) to compress raw ViT patch tokens before feeding them into the LLM.

We introduce \emph{Spatio-Temporal Token Scoring} (\textbf{STTS}), a lightweight plug-in module that is inserted into the ViT to selectively prune uninformative tokens before they propagate through the rest of the network. While we instantiate STTS on Molmo2 for all experiments, the module imposes no architecture-specific constraints, requiring only a standard ViT encoder and a token-to-LLM pathway -- both ubiquitous in modern VLMs~\cite{liu2023llava,wang2025internvl3,bai2025qwen3vltechnicalreport}.
At a high level, STTS operates in three coordinated steps: (1) a \emph{scorer} predicts the importance of each token along two complementary axes -- spatial saliency and inter-frame temporal redundancy; (2) a \emph{packing algorithm} converts the non-uniform, post-pruning sparse token sequences into compact dense tensors that yield genuine computational savings throughout the ViT; and (3) an \emph{auxiliary loss} provides an explicit training signal that guides the scorer to correctly identify temporally redundant regions. Since the pruning decision is made \emph{inside} the ViT, the reduced token count carries through to the LLM as well, achieving end-to-end efficiency gains across the entire VLM framework.

The following subsections describe each component in detail: Sections~\ref{sec: stts arch} and \ref{sec: stts bias injection} detail the scorer's design and spatial learning mechanism, Section~\ref{sec: stts rearrange} explains the packing algorithm, and Section~\ref{sec: stts aux loss} introduces the temporal auxiliary loss.

\begin{figure*}[t]
	\begin{center}
	\centering
        \includegraphics[width=\linewidth]{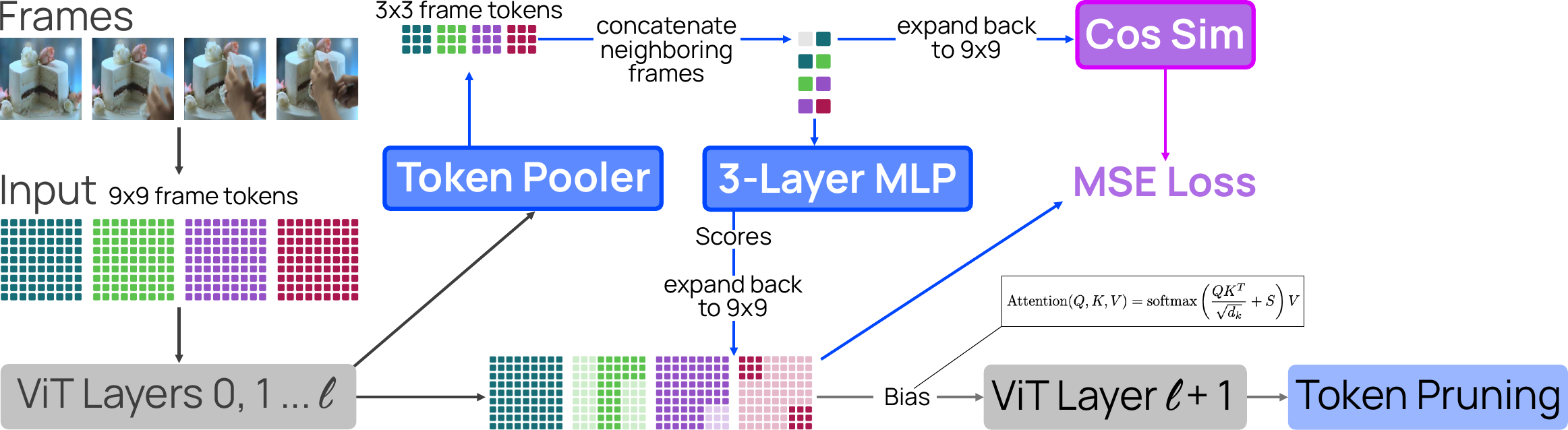}
        \caption{\textbf{Architectural and procedural overview of STTS.}  We use 9x9 tokens per frame for illustration. Vision features after ViT layer $l$ are first downsampled via pooling then scored. The scores are injected as attention bias for layer $l+1$ before the pruning algorithm is applied to allow for spatial pruning. The scores are also aligned with neighboring-frame per-patch cosine similarity for temporal pruning.}
        
		\label{fig: arch}
	\end{center}
\end{figure*}

\subsection{Scorer Architecture}
\label{sec: stts arch}

To achieve the spatial and temporal scoring outlined above, STTS features a simple architecture: a self-attention layer for pooling (Token Pooler) followed by a 3-Layer MLP for scoring, as demonstrated in Figure~\ref{fig: arch}. We insert STTS after a predetermined ViT layer $l$. Given an input $X \in \mathbb{R}^{T \times N \times D}$, representing $T$ video frames, $N$ patches, and a hidden dimension $D$, the features are first passed through layers $0,1,...,l$ of the ViT.

Before being scored by the MLP, the features $X_l$ are pooled with width $w$ to reduce the spatial dimension from $N$ to $N/w^2$. As introduced earlier, we use $w=3$ to align with the Molmo2 backbone.
To provide temporal context, the scorer's input for each frame $t$ is the concatenation of its pooled features with the pooled features of the previous frame, $t-1$, resulting in an input shape of $\mathbb{R}^{T \times (N/w^2) \times 2D}$. For the first frame ($t=0$), we concatenate it with a zero-padding tensor; its scores are ignored during pruning, as it lacks a preceding frame for temporal comparison. We thus always keep every first video frame intact.

\subsection{Bias Injection for Spatial Scoring}
\label{sec: stts bias injection}

The scorer outputs a single score for each $N/w^2$ pooled patch, where a lower score indicates lower importance. To apply these scores back to the original resolution, we expand them to the original $N$ patch locations, assigning the same score to all patches within their corresponding $w \times w$ block. The logarithm of these expanded scores, denoted as $S$, is then injected as a bias into the attention matrix of the subsequent ViT layer $l+1$:

$$\text{Attention}(Q, K, V) = \text{softmax}\left(\frac{QK^T}{\sqrt{d_k}} + S\right)V$$

This bias injection makes STTS end-to-end trainable, as it allows gradients from the final task loss to propagate back and teach the scorer to identify spatially salient tokens within each frame (or across each pair of neighboring frames) without explicit text conditioning.

\subsection{Token Pruning and Packing}
\label{sec: stts rearrange}

\begin{figure*}[t]
    \centering
    \begin{subfigure}{0.48\textwidth}
        \centering
        \includegraphics[width=\linewidth]{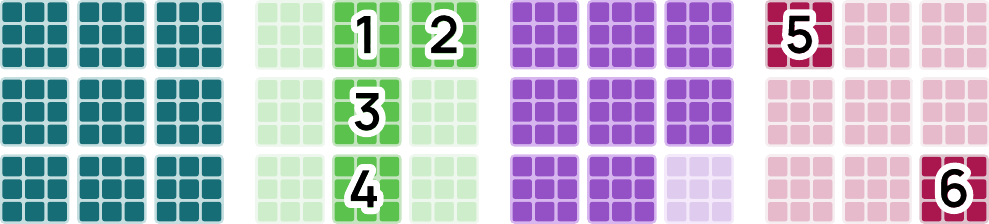}
        \caption{Before Pruning}
        \label{fig:before_pruning}
    \end{subfigure}\hfill
    \begin{subfigure}{0.48\textwidth}
        \centering
        \includegraphics[width=\linewidth]{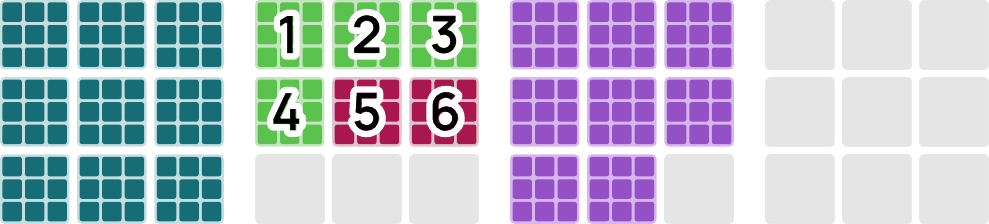}
        \caption{After Pruning}
        \label{fig:after_pruning}
    \end{subfigure}
    
    \caption{Visualization of the packing algorithm. (a) Before pruning, the scoring mechanism identifies the bottom-$k\%$ importance tokens (in dotted squares) to be removed. (b) To reduce tensor sparsity, the remaining tokens from Frame 2 (green) and Frame 4 (red) are consolidated into a single packed batch entry. Because Frame 1 is always untouched and Frame 3 retains high token counts, they remain independent.}
    \label{fig: packing}
\end{figure*}

Following layer $l+1$, we perform hard pruning by removing all tokens corresponding to the bottom-$k\%$ scores produced by STTS, where $k$ is a hyperparameter.

This introduces a critical challenge: our pruning is \textbf{video-aware} and inherently non-uniform across frames. While standard Image-ViTs process frames independently, our method may prune 80\% of tokens from a static frame (high redundancy) but only 10\% from a dynamic frame (high motion). This results in a sparse, ragged tensor. Because deep learning frameworks like PyTorch rely on dense, uniform tensors for efficient batched matrix multiplications, merely masking the pruned tokens yields no computational savings.

To overcome this and achieve actual hardware acceleration, we must pack the surviving tokens into a denser tensor. We treat the batch of frames $(T, N, D)$ as a set of $T$ variable-length token sequences. We then employ a \textit{first-fit descending} algorithm to pack these sparse sequences into a new, compact tensor of shape $(T', N, D)$, where $T' \le T$. The packing logic is summarized in 
Algorithm~\ref{alg:token_packing} of Appendix~\ref{sec:pseudocode}
and visualized in Figure~\ref{fig: packing}. We sort the frames by their valid token count (descending) and iterates through them, placing each frame's tokens into the first available packed ``bin'' (new frame) with sufficient capacity. This minimizes the total number of packed frames, $T'$, thereby maximizing computational throughput. Although the algorithm has a theoretical time complexity of $\mathcal{O}(T^2)$, the overhead is negligible because $T\ll N$, a point further supported by the efficiency gains demonstrated in Section~\ref{sec: speed}. 

Crucially, we generate a corresponding attention mask for the packed tensor. This mask ensures that tokens attend only to other tokens originating from the same source frame, preserving the integrity of the self-attention mechanism.

\subsection{Auxiliary Loss for Temporal Scoring}
\label{sec: stts aux loss}

While the scorer is intrinsically provided with temporal context by concatenating the current and previous frame features (as described in Section~\ref{sec: stts arch}), we found that this architectural design alone is insufficient when optimized solely with the primary task loss. In preliminary experiments, the LLM seemed indifferent to fine-grained temporal redundancy.
This is also reflected in Table~\ref{tab:scorer vs Heuristic vs random results} where the ``no aux'' variant of STTS falls significantly behind in downstream task performance.

To provide an explicit signal, we use Neighboring-Frame Cosine Similarity. We take the features $X_l$ from layer $l$ and apply the same $w\times w$ pooling as the scorer. We then L2-normalize the pooled features and compute the cosine similarity for each corresponding patch $i$ between adjacent frames $t$ and $t+1$:

\begin{equation*}
\text{CosSim}\left(X_{l, t}^{(i)}, X_{l, t+1}^{(i)}\right) = \frac{X_{l, t}^{(i)} \cdot X_{l, t+1}^{(i)}}{\left|\left|X_{l, t}^{(i)}\right|\right|_2 \cdot \left|\left|X_{l, t+1}^{(i)}\right|\right|_2}
\end{equation*}

where $X_{l, t}^{(i)}$ is the normalized, pooled feature for the $i$-th patch of frame $t$. We optimize the scorer to minimize the difference between its predicted scores and one minus these ``ground truth'' temporal similarity scores via an MSE loss, resulting in the per-element loss function:

\begin{equation*}
\mathcal{L}_{\text{sim}}(t, i) = \left( S_{t}^{(i)} - \left(1 - \text{CosSim}\left(X_{l, t-1}^{(i)}, X_{l, t}^{(i)}\right)\right) \right)^2
\end{equation*}

where $S_{t}^{(i)}$ is the score for the $w\times w$ patch $i$ of frame $t$ from STTS. $\mathcal{L}_{\text{sim}}$ guides STTS such that a higher similarity/redundancy should correlate with a lower importance score. Again, we set $\mathcal{L}_{\text{sim}}(0, i)=0$ for all patches $i$ in frame 0 since we don't prune them. Thus, the final end-to-end training objective is the sum of the task loss and the average of the above MSE loss:

\begin{equation*}
\mathcal{L}=\mathcal{L}_\text{task}+\frac{w^2}{TN}\sum_{t=0}^{T-1}\sum_{i=0}^{N-1}\mathcal{L}_\text{sim}(t,i)
\end{equation*}

\section{Experiments}

In this section, we conduct exhaustive experiments to demonstrate the effectiveness and soundness of STTS. We first delineate our training recipe in Section~\ref{ssec: train recipe}, then evaluate the trained models on standard short and long video QA tasks in Section~\ref{ssec:video_results}. We dive deep into the quantifiable efficiency gains using STTS in Section~\ref{sec: speed}. We also demonstrate how STTS does not affect image-only performance in
Appendix~\ref{ssec: image results}.

\subsection{Training Recipe}
\label{ssec: train recipe}
For our main results, we adopt the training recipe, data mixture, and model architecture from Molmo2~\cite{clark2026molmo2openweightsdata} as this is a very recent SoTA model with open source code and data. Our model architecture consists of the SigLIP 2 So400M/14 384px Image ViT~\cite{tschannen2025siglip2multilingualvisionlanguage} connected to a Qwen3-4B LLM~\cite{yang2025qwen3technicalreport} via a connector module. 
Due to limited compute resources and to expedite experimentation, we train only on the video QA subset of their data mixture. We start from the same pretrained video captioner checkpoint as Molmo2 and finetune it for 6,250 steps with batch size 64. Though this does mean that the model sees about $1/3$ of the videos that Molmo2 saw, we demonstrate in Table~\ref{tab:video_benchmark_results} that the baseline model still outperforms strong baselines like Qwen3-VL-4B~\cite{qwen3technicalreport}, validating our assumption that training for fewer steps does not cause significant performance degradations.

For optimization, we employ a cosine learning rate schedule with 200 warmup steps, using differential learning rates of 1e-5 for the LLM, 5e-6 for the ViT and projector, and 1e-4 for our STTS module. We always use $l=3$, meaning we apply STTS right after the 3rd ViT layer. We also allow bidirectional attention across all vision tokens in the LLM.

Following Molmo2's pre-processing strategies (including the aforementioned 3x3 spatial pooling), 
we first attempt to sample videos at 2 FPS; if this results in more than 64 frames, we fall back to uniformly sampling 64 frames across the entire video. The final frame of the video is always included. We also use the same sequence packing configuration as Molmo2 that concatenates multiple samples into one longer sequence before feeding them to the LLM. We pack on average 2 samples per batch, resulting in an effective batch size of 128.

\begin{table*}[t]
    \renewcommand{\arraystretch}{0.98}
    \centering
    \caption{\textbf{Effect of pruning strength on video benchmarks.} + STTS $k\%$ means $k\%$ pruning.
    Values that improve upon or remain within 0.5 points of the baseline are \textbf{bolded} and values within 1.0 point are \underline{underlined}.
    }
    \resizebox{\textwidth}{!}{
    \begin{tabular}{@{}l>{\columncolor{tableyellow!50}}c
    >{\columncolor{tableyellow!50}}c
    >{\columncolor{tableyellow!50}}c
    >{\columncolor{tableyellow!50}}c
    >{\columncolor{tableyellow!50}}c
    >{\columncolor{tableyellow!50}}c
    >{\columncolor{tablegreen!10}}c
    >{\columncolor{tablegreen!10}}c
    >{\columncolor{tablegreen!10}}c
    >{\columncolor{tablegreen!10}}c
    >{\columncolor{tablegreen!10}}c
    >{\columncolor{tablegreen!10}}c
    >{\columncolor{tablegreen!10}}c
    >{\columncolor{tableyellow!50}}c
    >{\columncolor{tablegreen!10}}c
    >{\columncolor{tableblue!10}}crr@{}}
        \textbf{Model} & 
        \mycell{NextQA}{test~\cite{nextqa}}
        & \mycell{Perception-Test}{test~\cite{perception_test}}
        & \mycell{MVBench}{test~\cite{mvbench}}
        & \mycell{Tomato}{test~\cite{tomato}}
        & \mycell{MotionBench}{val~\cite{motionbench}}
        & \mycell{Temp-Compass}{test~\cite{tempcompass}}
        & \mycell{VideoMME}{test~\cite{videomme}}
        & \mycell{VideoMME-Sub}{test~\cite{videomme}}
        & \mycell{LongVideo}{val~\cite{longvideobench}}
        & \mycell{LongVideo-Sub}{val~\cite{longvideobench}}
        & \mycell{MLVU}{val MCQ~\cite{mlvu}}
        & \mycell{LVBench}{test~\cite{lvbench}}
        & \mycell{VideoEvalPro}{test~\cite{videoevalpro}}
        & \newcell{Short avg.}
        & \newcell{Long avg.}
        & \newcell{Average}\\
        \midrule
        \color{gray} Qwen3-VL-4B~\cite{qwen3technicalreport} & \color{gray} 81.4 & \color{gray} 70.7 & \color{gray} 68.9 & \color{gray} 31.8 & \color{gray} 58.6 & \color{gray} 70.8 & \color{gray} 69.3 & \color{gray} 74.0 & \color{gray} 62.8 & \color{gray} - & \color{gray} 58.4 & \color{gray} 56.2 & \color{gray} 49.8 & \color{gray} 63.7 & \color{gray} 61.8 & \color{gray} 62.7 \\
        \color{gray} PLM-8B~\cite{cho2025PerceptionLM} & \color{gray} 84.1 & \color{gray} 82.7 & \color{gray} 77.1 & \color{gray} 33.2 & \color{gray} 61.4 & \color{gray} 72.7 & \color{gray} 58.3 & \color{gray} 65.4 & \color{gray} 56.9 & \color{gray} - & \color{gray} 52.6 & \color{gray} 44.5 & \color{gray} 47.2 & \color{gray} 68.5 & \color{gray} 54.2 & \color{gray} 61.3 \\
        \color{gray} InternVL3.5-8B~\cite{wang2025internvl3} & \color{gray} 81.7 & \color{gray} 72.7 & \color{gray} 72.1 & \color{gray} 24.6 & \color{gray} 56.6 & \color{gray} 70.3 & \color{gray} 66.0 & \color{gray} 68.6 & \color{gray} 62.1 & \color{gray} - & \color{gray} 53.2 & \color{gray} 43.4 & \color{gray} 48.1 & \color{gray} 63.0 & \color{gray} 56.9 & \color{gray} 60.0 \\
        \midrule
        
        Baseline (ours) & 83.9 & 78.7 & 72.6 & 36.5 & 61.0 & 69.9 & 62.8 & 67.6 & 61.5 & 60.9 & 70.3 & 42.0 & 47.6 & 67.0 & 59.0 & 63.0\\
        \midrule
        \color{molmocolor}+ STTS 30\% & \textbf{84.1} & \textbf{79.0} & \textbf{72.7} & \underline{35.6} & 59.2 & \textbf{69.6} & \textbf{63.4} & \textbf{68.5} & \textbf{61.1} & 59.2 & \underline{69.5} & \textbf{42.6} & \textbf{47.7} & \textbf{66.7} & \textbf{58.9} & \textbf{62.8}\\
        \color{molmocolor}+ STTS 40\% & \textbf{83.6} & 77.3 & \underline{71.8} & 34.6 & 59.2 & \underline{69.3} & \textbf{62.4} & \textbf{67.4} & \textbf{61.4} & \underline{60.2} & 67.5 & \underline{41.1} & \textbf{47.2} & \underline{66.0} & \underline{58.2} & \underline{62.1}\\
        \color{molmocolor}+ STTS 50\% & \textbf{83.7} & \underline{77.7} & \textbf{72.4} & 35.1 & 58.2 & \underline{69.2} & \textbf{62.4} & \textbf{67.2} & \textbf{61.0} & \underline{60.1} & 68.4 & 40.5 & 46.0 & \underline{66.1} & \underline{58.4} & \underline{62.3}\\
        
    \end{tabular}
    }
    \label{tab:video_benchmark_results}
\end{table*}

\subsection{Video Results}
\label{ssec:video_results}

We evaluate the efficacy of STTS by analyzing the trade-off between token reduction and model performance across a comprehensive suite of video benchmarks (Table~\ref{tab:video_benchmark_results}) as follows:

\noindent\textbf{Performance at 30\% Pruning.} 
We identify a ``sweet spot'' at 30\% pruning, where the model maintains or even exceeds baseline performance (e.g., on \textit{NextQA} and \textit{VideoMME}). This gain is a direct result of our scorer's learned capability. By utilizing downstream gradients, the scorer identifies and preserves ``task-essential'' tokens while the cosine similarity component effectively targets redundant background information. At 30\%, this synergistic filtering removes noise that would otherwise distract the attention mechanism, resulting in a set of tokens that are fewer in number but more effective for reasoning.

\noindent\textbf{Robustness at Higher Pruning Rates.} 
The method demonstrates remarkable robustness even under aggressive pruning regimes. At 50\% pruning—where half of the visual context is discarded—the model exhibits a minimal average performance decline of only 0.7\%. This stability is consistent across diverse tasks; for instance, on the comprehensive \textit{VideoMME} benchmark, performance dips by a mere 0.4 points. We attribute this broad resilience to the dual nature of our scorer: spatially, STTS learns to prioritize the semantic ``anchor'' tokens essential for reasoning; temporally, STTS safely discards the high volume of redundant temporal frames common in video data. Consequently, even with 50\% fewer tokens, the information density of the retained input remains sufficient.

\noindent\textbf{Non-Monotonic Behavior (40\% vs. 50\%).} 
We observe an intriguing trend where 50\% pruning (62.3 avg) outperforms 40\% pruning (62.1 avg). We attribute this to the interplay between the scorer's two objectives. At the intermediate 40\% level, the budget allows for tokens that are ``borderline''—not temporally redundant enough yet also lacking strong gradient support from the LLM. These tokens effectively act as noise, diluting the attention density. However, the more aggressive 50\% setup learns to identify these non-informative tokens and maximizes the signal-to-noise ratio of the visual input by pruning them.

\subsection{Efficiency Gains}
\label{sec: speed}

\begin{figure}[t]
    \centering
    \includegraphics[width=\linewidth]{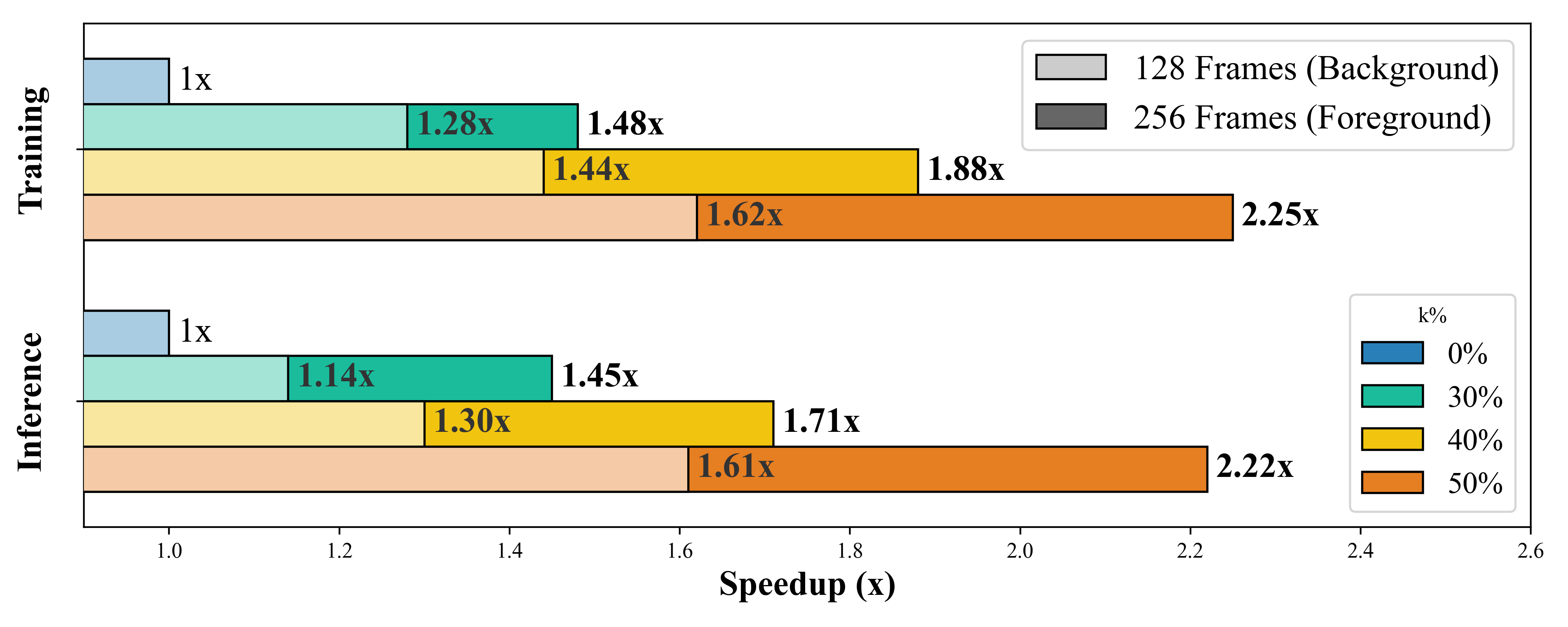}
    \caption{\textbf{Comparison of efficiency gains during training and inference across different pruning ratios ($k$)}. As $k$ increases, speedups greatly increase and become significantly larger when sampling more frames. See Supp. Sec. B for more details.}
    \label{fig:efficiency}
\end{figure}

Figure~\ref{fig:efficiency} quantifies the efficiency gains achieved by STTS across both training and inference phases. We include detailed throughput tables in
Appendix~\ref{sec:detailed thruput}.
To isolate computational performance from inter-node communication overhead, we conducted all profiling on a single node equipped with 8 H100 GPUs. Each padded training example consists of visual tokens (81 per frame for the baseline) combined with a maximum of 2048 text tokens. We evaluate performance under two settings: a 128-frame setup (64 frames, batch size 2), which matches our primary experimental configuration, and a more intensive 256-frame setup (256 frames, batch size 1). The latter represents a memory-constrained scenario where the unpruned baseline approaches the hardware's VRAM limits. We do not use sequence packing in the LLM in these experiments to ensure batch size consistency.

As illustrated in Figure~\ref{fig:efficiency}, increasing the pruning parameter $k$ consistently increases throughput for both training and inference. In the 128-frame setting, increasing $k$ to 50\% reduces the token load by approximately 33\%, yielding a \textbf{1.62x} speedup during training, while evaluation throughput on the MLVU benchmark—a characteristic benchmark for long video understanding—follows a nearly identical trajectory, achieving a \textbf{1.61x} speedup. 

Crucially, the computational benefits of STTS scale favorably with sequence length. In the 256-frame regime, the same 50\% pruning setting yields significantly larger speedups of \textbf{2.25x} for training and \textbf{2.22x} for inference. This disproportionate gain aligns with the quadratic $\mathcal{O}(N^2)$ complexity of the Transformer attention mechanism; as sequence length grows, computational savings from STTS become increasingly pronounced. The consistency between training and inference speedups confirms that the reduction in token processing overhead is robust across operational modes. This makes STTS particularly advantageous for deployment in memory-constrained environments or latency-sensitive applications requiring long-context video understanding.

Finally, we observe a marginal attenuation in relative speedup during inference compared to training. We attribute this to the training pipeline's use of \texttt{torch.compile}. STTS plays nicely with static graph execution; because all examples are padded to a uniform sequence length, the static graph maximizes the relative computational gains brought by token reduction. In contrast, inference loops handle dynamic sequence lengths during prefill, resulting in slightly different overhead characteristics.

\section{Ablation Studies}
To justify STTS's design and hyperparameter selection, we conduct an extensive number of ablation studies to demonstrate STTS's novelty and necessity. First, we compare our learned scorer against a non-learnable heuristic baseline in Section~\ref{sec: scorer vs easy ablation}. Next, we ablate the choice of the ViT injection layer depth ($l$) in Section~\ref{ssec:layer_depth}. We then explore the benefits of Test-Time Scaling (TTS) on long video benchmarks in Section~\ref{ssec:tts}. Finally, we visualize and analyze the pruning behavior of our scorer in Section~\ref{ssec:scorer_behavior}. We further compare STTS with ViT-only pruning methods in
Appendix~\ref{ssec:vit_only_comparison}.

\subsection{Scorer Pruning vs. Heuristic Pruning}
\label{sec: scorer vs easy ablation}
As described in Section~\ref{sec: stts aux loss}, we use neighboring-frame cosine similarity to guide STTS. A natural baseline, therefore, is to bypass the learned scorer and use this similarity signal directly. This ``heuristic pruning'' approach involves sorting the computed similarities and pruning the top-$k\%$ of visual tokens from neighboring frames that are most similar.
We also include results from a model trained with STTS without employing the auxiliary loss.
Finally, we include $k\%$ random pruning to establish a lower bound and contextualize the results.

\begin{figure*}[t] %
    \centering
    
    \begin{minipage}[c]{0.49\textwidth}
        \centering
        \captionof{table}{Comparison between different pruning methods using 50\% pruning. With Random as the baseline, STTS outperforms Heuristic, especially on long videos.}
        \begin{tabular}{@{}l
        >{\columncolor{tableyellow!50}}c
        >{\columncolor{tablegreen!10}}c
        >{\columncolor{tableblue!10}}crr@{}}
            \textbf{Method\;}
            & \newcell{Short avg.}
            & \newcell{Long avg.}
            & \newcell{Average}\\
            \midrule
            Random & 65.3 & 57.5 & 61.4\\
            Heuristic & 66.0 & 57.9 & 62.0 \\
            STTS (No Aux) & 64.4 & 55.5 & 60.0\\
            \textcolor{molmocolor}{STTS} & \textcolor{molmocolor}{66.1} & \textcolor{molmocolor}{58.4} & \textcolor{molmocolor}{62.3} \\
        \end{tabular}
        \label{tab:scorer vs Heuristic vs random results}
    \end{minipage}\hfill
    \begin{minipage}[c]{0.49\textwidth}
        \centering
        \includegraphics[width=0.8\linewidth]{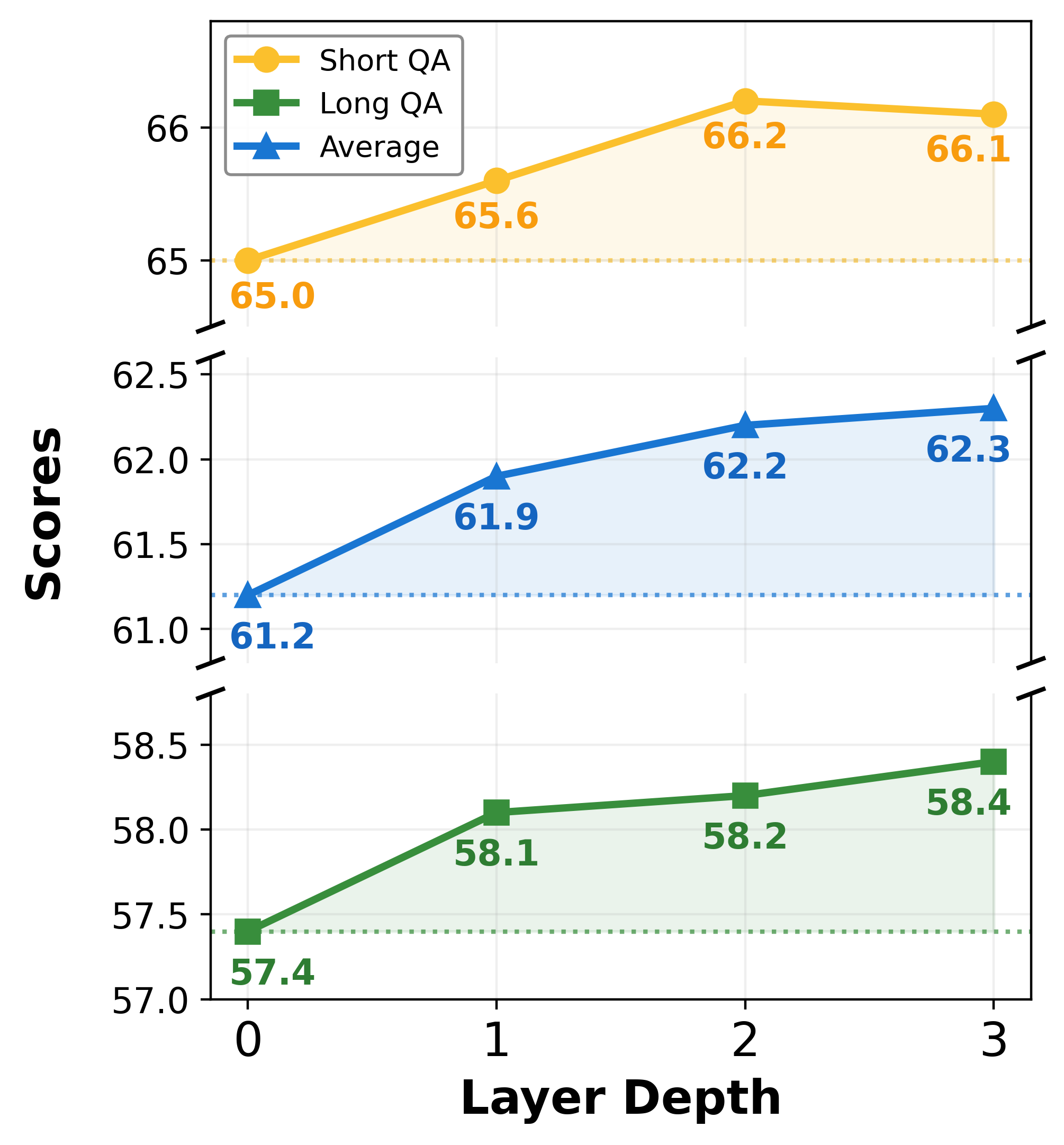}
        \vspace{-1em}
        \captionof{figure}{$l=0,1$ hurts performance, while $l=2$ is marginally weaker than $l=3$.}
        \label{fig:depth_ablation}
    \end{minipage}
\vspace{-1em}
\end{figure*}

Table~\ref{tab:scorer vs Heuristic vs random results} demonstrates that random pruning underperforms both heuristic and scorer-based methods by a significant margin of approximately 1\%.
The ``no aux'' variant of STTS performs even worse than Random, validating our assumption that the VLM backbone itself is indifferent towards temporal redundancy and cannot provide good pruning signals on its own.
While the scorer performs marginally better than the heuristic on short videos, it extends this lead to 0.5\% on long videos. Since long video QA benchmarks typically consist of hour-long sequences, FPS-based sampling would far exceed our 64-frame budget (Section \ref{ssec: train recipe}), falling back to uniform sampling. The resulting sparse frame selection leaves minimal temporal redundancy across frames. In this context, the scorer effectively distinguishes salient tokens by leveraging spatial signals to compensate for weak temporal cues, thereby maintaining both efficiency and performance.

\subsection{Selecting Pruning Layer Depth ($l$)}
\label{ssec:layer_depth}
The injection layer $l$ is a crucial hyperparameter, as ViT layers serve different functions. Early layers (e.g., 0-4) are thought to handle low-level feature extraction and token contextualization, while deeper layers (e.g., 12-16) aggregate more complex semantic information. We hypothesized that pruning too early (e.g., $l=0$) would prevent the ViT from forming robust patch representations before critical information was discarded. To validate this hypothesis, we ablate this choice by training four separate models with $l \in \{0, 1, 2, 3\}$.

Figure~\ref{fig:depth_ablation} illustrates a positive correlation between performance and depth $l$. The significant 1\% performance gap between $l=0$ and $l=3$, alongside the 0.5\% gap between $l=1$ and $l=3$, indicates that premature pruning is detrimental. We hypothesize that this performance degradation arises either because the scorer lacks sufficient contextualized information to identify salient tokens or because bias injection and hard pruning disproportionately damage the ViT's initial, more sensitive layers. Since performance at $l=2$ is only marginally inferior to $l=3$, we do not evaluate deeper layers; increasing $l$ further would diminish the computational efficiency gains derived from token pruning. Consequently, these findings justify our selection of $l=3$ for all subsequent experiments.

\subsection{Test-Time Scaling for Long Video Benchmarks}
\label{ssec:tts}

\begin{table*}[t]
    \renewcommand{\arraystretch}{0.98}
    \centering
    \caption{\textbf{Impact of Test-Time Scaling (TTS) on Long Video Benchmarks.} Performance comparison when increasing number of frames sampled (\# Fr) \textbf{only during inference}. Significant improvements over the baseline (0\%) are \textbf{bolded}.}
    \resizebox{0.6\linewidth}{!}{
    \begin{tabular}{@{}lcc>{\columncolor{tablegreen!10}}c
    >{\columncolor{tablegreen!10}}c
    >{\columncolor{tablegreen!10}}c
    >{\columncolor{tablegreen!10}}c
    >{\columncolor{tablegreen!10}}c
    >{\columncolor{tablegreen!10}}c
    >{\columncolor{tablegreen!10}}c
    >{\columncolor{tableblue!10}}c@{}}
        \textbf{$k\%$} & TTS\; & \# Fr\; &
        \mycell{VideoMME}{test~\cite{videomme}}
        & \mycell{VideoMME-Sub}{test~\cite{videomme}}
        & \mycell{LongVideo}{val~\cite{longvideobench}}
        & \mycell{LongVideo-Sub}{val~\cite{longvideobench}}
        & \mycell{MLVU}{val MCQ~\cite{mlvu}}
        & \mycell{LVBench}{test~\cite{lvbench}}
        & \mycell{VideoEvalPro}{test~\cite{videoevalpro}}
        & \newcell{Long avg.}\\
        \midrule
        
        0\% & \xmark & 64 & 62.8 & 67.6 & 61.5 & 60.9 & 70.3 & 42.0 & 47.6 & 59.0\\
        \midrule
        \multirow{2}{*}{30\%} & \xmark & 64 & 63.4 & 68.5 & 61.1 & 59.2 & 69.5 & 42.6 & 47.7 & 58.9\\
         & \cmark & 92 & 62.9 & \textbf{69.1} & \textbf{62.7} & 60.8 & 70.5 & \textbf{44.9} & \textbf{49.6} & \textbf{60.1}\\
         \midrule
        \multirow{2}{*}{40\%} & \xmark & 64 & 62.4 & 67.4 & 61.4 & 60.2 & 67.5 & 41.1 & 47.2 & 58.2\\
         & \cmark & 107 & 62.0 & \textbf{68.1} & \textbf{62.7} & 60.6 & 68.6 & \textbf{42.5} & \textbf{48.6} & 59.0\\
        \midrule
        \multirow{2}{*}{50\%} & \xmark & 64 & 62.4 & 67.2 & 61.0 & 60.1 & 68.4 & 40.5 & 46.0 & 58.4\\
         & \cmark & 128 & 62.8 & \textbf{69.0} & 60.9 & 59.9 & 69.2 & \textbf{44.7} & \textbf{49.3} & \textbf{59.4}\\
        
    \end{tabular}
    }
    \label{tab:tts_long_video_benchmark_results}
\end{table*}
In this section, we analyze the impact of our pruning method combined with Test-Time Scaling (TTS) on long video understanding. Pruning tokens during inference reduces the computational load per frame; for instance, pruning 50\% of tokens from 64 frames results in a visual token count equivalent to only 32 unpruned frames. To ensure a fair comparison and fully utilize the available token budget, we employ TTS on models trained with $k\%$-pruning on 64 frames by increasing the frame count proportionally (e.g., increasing to 128 frames for the 50\% pruning setting) to match the baseline's visual token usage. We note that we do \textbf{not} retrain any model; we only increase frame count during inference.

As shown in Table~\ref{tab:tts_long_video_benchmark_results}, we observe steady performance improvements across all TTS configurations compared to the baseline. Specifically, \textbf{30\% + TTS} achieves a Long QA average of \textbf{60.1}, outperforming the baseline by a significant \textbf{1.1\%} margin. Similarly, despite the aggressive pruning rate, \textbf{50\% + TTS} achieves an average of \textbf{59.4}, surpassing the baseline by roughly \textbf{0.5\%}.

Furthermore, comparing the pruned models with and without scaling reveals the efficacy of this approach. All TTS methods consistently outperform their pre-TTS counterparts by a margin of roughly \textbf{1\%}. This indicates that STTS effectively trades off spatial redundancy for temporal density: by pruning less informative tokens, we can process a significantly larger number of frames (up to 128) within the same computational envelope, thereby capturing richer temporal context essential for long video understanding.

\begin{figure*}[!t]
	\begin{center}
	\centering
        \includegraphics[width=\linewidth]{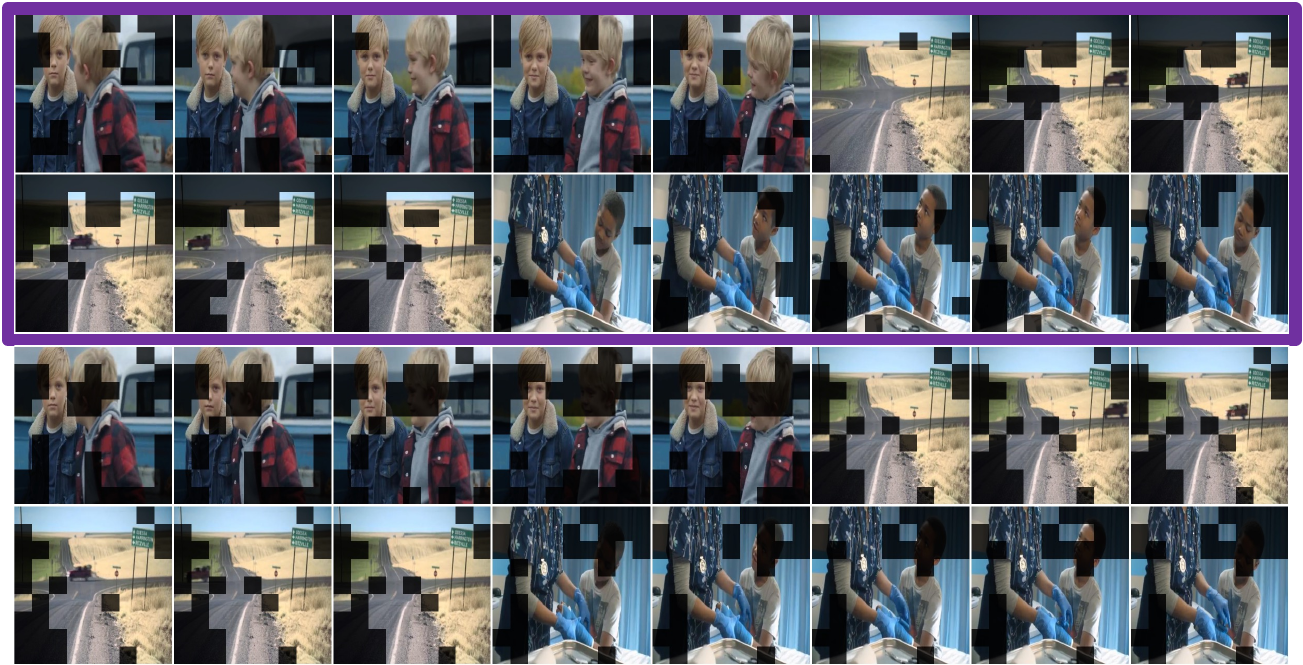}
        \vspace{-2em}
		\caption{Visualizations of STTS (purple box) vs. the heuristic. STTS attempts to keep important information despite temporal redundancy, while the heuristic prunes them regardless due to the lack of knowledge.}
		\label{fig: vis}
	\end{center}
\end{figure*}

\subsection{Analyzing Scorer Behavior}
\label{ssec:scorer_behavior}
Figures~\ref{fig:teaser} and~\ref{fig: vis} visualize the token pruning results of our STTS scorer on selected Molmo2-Caption~\cite{clark2026molmo2openweightsdata} examples. We present two distinct scenarios to highlight the differences in pruning behavior. In both examples, the upper purple boxes illustrate the spatial patches pruned per frame using STTS, whereas the lower green boxes display the results from the non-learnable heuristic baseline.

The first example features a 2D platformer game (similar to \textit{Super Mario}), characterized by a static background and a highly dynamic foreground where platforms and characters continuously move. Because a large portion of the background remains identical across frames, the heuristic method struggles to distinguish semantic content; instead, it blindly---and seemingly randomly---prunes redundant tokens based solely on simple inter-frame similarities. Conversely, STTS exhibits a highly interpretable and logical pruning pattern. Rather than treating all visually similar patches equally, STTS prioritizes the retention of foreground elements. Because the STTS scorer takes rich visual tokens as input and is optimized via downstream gradients from the LLM, it learns to recognize that foreground objects hold greater semantic importance. Consequently, STTS aggressively prunes the static background while consistently preserving tokens corresponding to the player, moving objects, and active platforms.

The second example showcases a real-life video sequence, further underscoring the limitations of a rigid, non-learnable algorithm compared to STTS. The heuristic method erroneously prunes away faces despite significant yet small changes in posture and facial expression. We hypothesize that this failure occurs because the heuristic relies on raw feature similarity applied at a shallow layer ($l=3$). By this stage, attention mechanisms have not blended enough contextual information across tokens, and the fine-grained visual differences of the facial movements might not be well-encoded, causing the heuristic to mistake them for being simply "faces" and thus redundant. STTS, however, implicitly understands the semantic weight of human faces and expressions in video reasoning tasks and recognizes these details as critical narrative elements and preserves them entirely, ensuring no loss of important information.

In conclusion, these visualizations validate that our learnable STTS approach significantly outperforms static heuristic methods. While heuristic approaches blindly discard tokens based on superficial feature similarities, STTS acts as an intelligent semantic filter. Guided by the downstream LLM, it effectively distinguishes between functionally irrelevant backgrounds and critical foreground dynamics, yielding a highly efficient yet expressive token representation for complex video understanding.

\subsection{Analysis of Performance Degradation}
\label{sec:performance_analysis}

To rigorously evaluate the robustness of STTS under strict computational constraints, we analyze the performance degradation as the vision token pruning ratio $k$ increases. Table~\ref{tab:k_effect} in Appendix~\ref{sec: detailed degrade} and the right subfigure of Figure~\ref{fig:teaser} illustrate the impact of aggressively pruning visual tokens on the model's Question Answering (QA) capabilities.

Crucially, to properly contextualize these results, we first establish the true baseline performance by examining the extreme case of $k=100$. At this setting, 100\% of the vision tokens are pruned, providing no visual information to the model and reducing the task to pure text-based reasoning. The Random method achieves a QA Average of 44.6\% (46.6\% for Short QA and 42.5\% for Long QA). This demonstrates that nearly 45\% of the questions can be correctly answered by relying solely on linguistic priors and inherent dataset biases, without requiring any actual visual context. Consequently, any performance gains achieved above this $\sim$45\% floor represent genuine, visually-grounded multimodal reasoning rather than mere language exploitation.

Viewed through this lens, the improvements yielded by STTS are highly significant. While both methods naturally experience a decline in performance as the token pruning ratio increases, STTS degrades at a demonstrably slower and flatter rate compared to random token dropping. 

From the outset at $k=50$ (where 50\% of the vision tokens are discarded), STTS establishes a clear advantage over the Random baseline. As the token budget becomes increasingly constrained, this performance gap widens substantially. For instance, at a severe pruning ratio of $k=80$ (retaining only 20\% of the vision tokens), STTS achieves a QA Average of 59.8\% compared to Random's 57.5\%---a 2.3\% absolute improvement. Given that the effective range for visually-driven performance is heavily compressed by the 45\% text-only baseline, these consistent gains highlight the core strength of our approach. By intelligently scoring and preserving the most informative spatio-temporal tokens, STTS ensures robust multimodal grounding even when operating under extreme token reduction constraints.

\section{Conclusion}
\label{sec:conclusion}

In this work, we introduced Spatio-Temporal Token Scoring (STTS), an end-to-end trainable framework that unifies token pruning across both the vision encoder and the LLM. By leveraging downstream task gradients alongside an auxiliary temporal loss, STTS effectively filters redundant background noise while preserving critical semantic foregrounds—eliminating the need for complex, text-conditioned merging. Our experiments confirm that STTS safely reduces visual token counts by 50\%, accelerating both training and inference by over 60\% with negligible performance degradation across 13 diverse video QA benchmarks. Furthermore, we demonstrated that STTS pairs naturally with test-time scaling, unlocking the ability to process substantially longer temporal contexts under strict computational constraints. Ultimately, STTS offers a simple and highly interpretable solution to the VLM efficiency bottleneck, paving the way for more accessible and scalable video understanding systems.

\section*{Acknowledgments}
This work would not be possible without the support of our colleagues at Ai2, in particular the PRIOR team. We thank Mohammadreza Salehi for discussing test-time scaling applications of STTS for long video evaluations. We thank other members of the PRIOR team for providing advice and feedback on various aspects of the designs of STTS.

This work was supported in part by NSF IIS2404180.

\bibliographystyle{abbrvnat}
\bibliography{eccv/main}

\clearpage

\appendix
\section*{Appendix}
\begin{table*}[!ht]
    \renewcommand{\arraystretch}{0.98}
    \centering
    \caption{\textbf{Image benchmark results} between a different version of Molmo2 trained with slightly different data and our STTS variant. In this attempt, both models are trained on the exact same full mixture with both video and image data. We can see that even though our model applies pruning on videos, the overall image performance did not degrade.}
    \setlength{\tabcolsep}{3pt}
    \resizebox{\textwidth}{!}{
    \begin{tabular}{@{}l>{\columncolor{tableyellow!50}}c
    >{\columncolor{tableyellow!50}}c
    >{\columncolor{tableyellow!50}}c
    >{\columncolor{tableyellow!50}}c
    >{\columncolor{tableyellow!50}}c
    >{\columncolor{tableyellow!50}}c
    >{\columncolor{tableyellow!50}}c
    >{\columncolor{tableyellow!50}}c
    >{\columncolor{tableyellow!50}}c
    >{\columncolor{tableyellow!50}}c
    >{\columncolor{tableyellow!50}}c
    >{\columncolor{tablegreen!10}}c
    >{\columncolor{tablegreen!10}}c
    >{\columncolor{tableyellow!50}}c
    >{\columncolor{tablegreen!10}}c
    >{\columncolor{tableblue!10}}c}
        \textbf{$k\%$} & 
        \mycell{AI2D}{test~\cite{ai2_diagram}}
        & \mycell{ChartQA}{test~\cite{chartqa}}
        & \mycell{DocVQA}{test~\cite{mathew2021docvqa}}
        & \mycell{InfoQA}{test~\cite{infoqa}}
        & \mycell{TextVQA}{val~\cite{textqa}}
        & \mycell{VQA v2.0}{val ~\cite{goyal2017making}}
        & \mycell{RWQA}{\cite{realworldqa}}
        & \mycell{MMMU}{val~\cite{yue2024mmmu}}
        & \mycell{MathVista}{testmini~\cite{lu2024mathvista}}
        & \mycell{CountBench}{~\cite{beyer2024paligemma}}
        & \mycell{PixMoCount}{test \cite{molmov1}}
        & \mycell{MuirBench}{\cite{wang2024muirbench}}
        & \mycell{MMIU}{\cite{meng2024mmiumultimodalmultiimageunderstanding}}
        & \newcell{Img avg.}
        & \newcell{MultiImg avg.}
        & \newcell{Average}\\
        \midrule
        0\% & 95.0 & 84.1 & 91.9 & 77.2 & 85.5 & 86.0 & 74.6 & 49.6 & 57.4 & 94.3 & 91.1 & 61.2 & 54.2 & 80.6 & 57.7 & 77.1 \\
        50\% & 94.3 & 83.9 & 91.3 & 77.3 & 85.7 & 86.2 & 75.4 & 50.0 & 57.4 & 94.9 & 89.6 & 62.0 & 54.9 & 80.5 & 58.4 & 77.3 \\
    \end{tabular}
    }%
    \label{tab:image_benchmark_results}
\end{table*}

\begin{table}[!ht]
    \centering
    \caption{Performance comparison at a 50\% pruning rate on video QA tasks. STTS significantly outperforms inference-only baselines (Spatial Heuristic, ToMe) and a fully trained ToMe model.}
    \label{tab:vit_only_comparison}
    
    \begin{tabular}{@{}l
    >{\columncolor{tableyellow!50}}c
    >{\columncolor{tablegreen!10}}c
    >{\columncolor{tableblue!10}}c@{}}
        \textbf{Method}
        & \newcell{Short avg.}
        & \newcell{Long avg.}
        & \newcell{Average}\\
        \midrule
        Heuristic [Inference Only] & 63.1 & 55.1 & 59.1\\
        ToMe [Inference Only] & 62.4 & 55.9 & 59.2\\
        ToMe & 65.6 & 56.6 & 61.1 \\
        \textcolor{molmocolor}{STTS} & \textcolor{molmocolor}{66.1} & \textcolor{molmocolor}{58.4} & \textcolor{molmocolor}{62.3}\\
    \end{tabular}
\end{table}

\section{Image Results}
\label{ssec: image results}

In Table~\ref{tab:image_benchmark_results}, we show a performance comparison between a different version of Molmo2 (trained on slightly different data) and the same model trained with STTS on image-QA benchmarks. This demonstrates how STTS can prune video tokens without harming image-only task accuracy. We attribute this to STTS's ability to use downstream gradients to learn an optimal token selection policy, ensuring that only non-essential tokens are removed. We surprisingly see a 1-point improvement on multi-image QA. We hypothesize this is a transfer learning effect: because both video analysis and multi-image QA require processing multiple visual frames simultaneously, the temporal reasoning skills the model learned from video data unexpectedly boosted its multi-image performance.

\section{Detailed Throughput Tables}
\label{sec:detailed thruput}

We include detailed information of our throughput analysis here, where Table~\ref{tab: speed} is for training, while Table~\ref{tab: eval speed} is for inference on MLVU.

\begin{table}[t]
    \centering
    \caption{Comparison of training speed between baseline and different pruning setup $k$'s. The number of tokens per instance decreases as $k$ increases, while throughput (batches per second) and speedup increases. Efficiency gains grew larger as we increase max number of frames sampled.}
    \label{tab: speed}
    \begin{tabular}{l|c|ccc}
        $k\%$ & \# Fr & \;\;Toks/Inst.\;\; & Throughput\;\; & Speedup\\
        \midrule
        0\% & \multirow{4}{*}{128} & 15670 & 0.1932 & 1x \\
        30\% & & 12560 & 0.2478 & 1.28x\\
        40\% & & 11524 & 0.2786 & \textbf{1.44x}\\
        50\% & & 10486 & 0.3130 & \textbf{1.62x}\\
        \midrule
        0\% & \multirow{4}{*}{256} & 25307 & 0.0549 & 1x\\
        30\% & & 19087 & 0.0811 & \textbf{1.48x}\\ %
        40\% & & 17013 & 0.0977 & \textbf{1.88x} \\
        50\% & & 14939 & 0.1233 & \textbf{2.25x}
    \end{tabular}
\end{table}

\begin{table}[t]
    \centering
    \caption{Comparison of inference speed between baseline and different pruning setup $k$'s on MLVU. We observe identical trends during training.}
    \label{tab: eval speed}
    \begin{tabular}{l|c|ccc}
        $k\%$ & \# Fr  & \;\;Throughput\;\; & Speedup\\
        \midrule
        0\% & \multirow{4}{*}{128} & 1.0186 & 1x \\
        30\% & & 1.1651 & 1.14x\\
        40\% & & 1.3270 & 1.30x\\
        50\% & & 1.6439 & \textbf{1.61x}\\
        \midrule
        0\% & \multirow{4}{*}{256} & 0.2641 & 1x\\
        30\% & & 0.3836 & \textbf{1.45x}\\ %
        40\% & & 0.4516 & \textbf{1.71x} \\
        50\% & & 0.5870 & \textbf{2.22x}
    \end{tabular}
\end{table}

\section{Comparison with ViT-Only Pruning Baselines}
\label{ssec:vit_only_comparison}

While STTS is uniquely designed to prune jointly across both the vision encoder and the language model, we benchmark its performance against established pruning baselines to isolate its architectural advantages. Specifically, we compare STTS against inference-only applications of the heuristic version of STTS and Token Merging (ToMe)~\cite{bolya2023tokenmergingvitfaster}, as well as a fully trained version of ToMe. To adapt ToMe for this architecture, we apply the merging method within the ViT and pass the modified, pooled patches to the LLM.

The results are shown in Table~\ref{tab:vit_only_comparison}. The substantial performance advantage of STTS over both inference-only baselines underscores a critical requirement: the model must be actively trained on pruned input sequences instead of relying on the complete set of 81 tokens per frame. Furthermore, STTS outperforms the fully trained ToMe baseline. Because ToMe and similar ViT-centric reduction methods are designed primarily for image-level tasks, applying them to video streams merges tokens without sufficient structural or temporal awareness. This often compromises the fine-grained details required for complex video reasoning. In contrast, by learning directly from downstream video task objectives, STTS effectively captures both the spatial and temporal importance of every token across frames.

Quantitatively, at a 50\% pruning rate, while introducing training to ToMe improves its performance over its inference-only counterpart, it still suffers from this inherent limitation. STTS achieves a QA Average of 62.3, mitigating the roughly 1-point performance drop seen in the trained ToMe baseline across both Short QA and Long QA tasks. This substantial margin demonstrates that STTS avoids the pitfalls of naive, image-based pruning, relying instead on a superior cross-modal strategy that preserves critical, fine-grained spatiotemporal information.

\begin{algorithm}[t]
\caption{Token Packing via First-Fit Descending}
\label{alg:token_packing}
\begin{algorithmic}[1]
\Require Input tensor $X \in \mathbb{R}^{T \times N \times D}$
\Require Valid mask $M \in \{0, 1\}^{T \times N}$
\State $C_{\text{valid}} \gets \text{count\_valid\_tokens}(M, \text{dim}=1)$ \Comment{Shape: $(T,)$}
\State $I_{\text{sorted}} \gets \text{argsort\_descending}(C_{\text{valid}})$
\State $P_{\text{load}} \gets \text{zeros}(T)$ \Comment{Token load per packed frame}
\State $P_{\text{assign}} \gets \text{zeros}(T, \text{dtype=int})$ \Comment{Map old frame $i$ to new frame $j$}
\State $P_{\text{offset}} \gets \text{zeros}(T, \text{dtype=int})$ \Comment{Start pos. of frame $i$ in new frame $j$}
\State
\For{$i$ in $I_{\text{sorted}}$}
    \State $count \gets C_{\text{valid}}[i]$
    \State $j \gets \text{find\_first\_fit}(count, P_{\text{load}}, N)$ \Comment{Find first new frame $j$ that fits $count$ tokens}
    \State $P_{\text{assign}}[i] \gets j$
    \State $P_{\text{offset}}[i] \gets P_{\text{load}}[j]$
    \State $P_{\text{load}}[j] \gets P_{\text{load}}[j] + count$
\EndFor
\State
\State $T_{\text{packed}} \gets \text{num\_non\_empty\_frames}(P_{\text{load}})$
\State $X_{\text{packed}} \gets \text{zeros}(T_{\text{packed}}, N, D)$
\State $\text{Mask}_{\text{packed}} \gets \text{zeros}(T_{\text{packed}}, N, N)$
\State \Comment{Scatter tokens into new tensor based on assignment}
\State $\text{scatter\_tokens}(X, M, P_{\text{assign}}, P_{\text{offset}}, \text{out=}X_{\text{packed}})$
\State \Comment{Build block-diagonal mask for packed tensor}
\State $\text{build\_attention\_mask}(P_{\text{assign}}, P_{\text{offset}}, C_{\text{valid}}, \text{out=}\text{Mask}_{\text{packed}})$
\State \Return $X_{\text{packed}}, \text{Mask}_{\text{packed}}$
\end{algorithmic}
\end{algorithm}

\section{Pseudocode for STTS Packing Algorithm}
\label{sec:pseudocode}
In Algorithm~\ref{alg:token_packing}, we demonstrate the pseudocode that is used for packing sparse tokens into a denser tensor within the ViT. Our algorithm optimizes for maximal compression, while using an $\mathcal{O}(T^2)$ algorithm (from the for-loop and the method \texttt{find\_first\_fit}) to find the best bin for each frame.

\section{Detailed Performance Degradation Tables}
\label{sec: detailed degrade}
Here we include Table~\ref{tab:k_effect} to complement the right subfigure of Figure~\ref{fig:teaser}.

\begin{table}[!h]
    \centering
    \caption{Comparing numerical performance values between Random and STTS from $k=50$ to $k=90$. STTS consistently outperforms Random. Text-only baseline $k=100$ provided as lower bound.}
    \label{tab:k_effect}
    \begin{tabular}{@{}lc
        >{\columncolor{tableyellow!50}}c
        >{\columncolor{tablegreen!10}}c
        >{\columncolor{tableblue!10}}c@{}}
        \textbf{$k$} & \textbf{Method} 
        & \newcell{Short avg.} 
        & \newcell{Long avg.} 
        & \newcell{Average}\\
        \midrule
        \multirow{2}{*}{50} & Random & 65.3 & 57.5 & 61.4\\
           & \textcolor{molmocolor}{STTS} & \textcolor{molmocolor}{66.1} & \textcolor{molmocolor}{58.4} & \textcolor{molmocolor}{62.3} \\
        \midrule
        \multirow{2}{*}{60} & Random & 63.9 & 56.3 & 60.1\\
           & \textcolor{molmocolor}{STTS} & \textcolor{molmocolor}{65.3} & \textcolor{molmocolor}{57.9} & \textcolor{molmocolor}{61.6}\\
        \midrule
        \multirow{2}{*}{70} & Random & 63.2 & 55.0 & 59.1\\
           & \textcolor{molmocolor}{STTS} & \textcolor{molmocolor}{65.0} & \textcolor{molmocolor}{56.4} & \textcolor{molmocolor}{60.7}\\
        \midrule
        \multirow{2}{*}{80} & Random & 61.3 & 53.6 & 57.5\\
           & \textcolor{molmocolor}{STTS} & \textcolor{molmocolor}{63.7} & \textcolor{molmocolor}{55.8} & \textcolor{molmocolor}{59.8}\\
        \midrule
        \multirow{2}{*}{90} & Random & 58.5 & 51.0 & 54.8\\
           & \textcolor{molmocolor}{STTS} & \textcolor{molmocolor}{60.2} & \textcolor{molmocolor}{52.1} & \textcolor{molmocolor}{56.2} \\
        \midrule
        100 & N/A & 46.6 & 42.5 & 44.6\\
    \end{tabular}
\end{table}

\clearpage

\end{document}